\documentclass[a4paper, 10pt, conference]{ieeeconf}      
\usepackage{FG2021}
\usepackage[accsupp]{axessibility} 

\FGfinalcopy 

\IEEEoverridecommandlockouts                              
\usepackage{graphics} 
\usepackage{epsfig} 
\usepackage{amsmath} 

\def\FGPaperID{289} 

\title{\LARGE \bf
High-Accuracy RGB-D Face Recognition via Segmentation-Aware Face Depth Estimation and Mask-Guided Attention Network
}

\author{\parbox{16cm}{\centering
    {\large Meng-Tzu Chiu$^1$, Hsun-Ying Cheng$^1$, Chien-Yi Wang$^2$, and Shang-Hong Lai$^{1\text{,}2}$}\\
    {\normalsize
    $^1$ Department of Computer Science, National Tsing Hua University, Taiwan\\
    $^2$ Microsoft AI R\&D Center, Taiwan}}
}

\begin{document}

%
%
%




\IEEEoverridecommandlockouts\pubid{\makebox[\columnwidth]{978-1-6654-3176-7/21/\$31.00~\copyright{}2021 IEEE \hfill}
\hspace{\columnsep}\makebox[\columnwidth]{ }}

\ifFGfinal
\thispagestyle{empty}
\pagestyle{empty}
\else
\author{Anonymous FG2021 submission\\ Paper ID \FGPaperID \\}
\pagestyle{plain}
\fi
\maketitle

\begin{abstract}

Deep learning approaches have achieved highly accurate face recognition by training the models with very large face image datasets. Unlike the availability of large 2D face image datasets, there is a lack of large 3D face datasets available to the public. Existing public 3D face datasets were usually collected with few subjects, leading to the over-fitting problem. This paper proposes two CNN models to improve the RGB-D face recognition task. The first is a segmentation-aware depth estimation network, called DepthNet, which estimates depth maps from RGB face images by including semantic segmentation information for more accurate face region localization. The other is a novel mask-guided RGB-D face recognition model that contains an RGB recognition branch, a depth map recognition branch, and an auxiliary segmentation mask branch with a spatial attention module. Our DepthNet is used to augment a large 2D face image dataset to a large RGB-D face dataset, which is used for training an accurate RGB-D face recognition model. Furthermore, the proposed mask-guided RGB-D face recognition model can fully exploit the depth map and segmentation mask information and is more robust against pose variation than previous methods. Our experimental results show that DepthNet can produce more reliable depth maps from face images with the segmentation mask. Our mask-guided face recognition model outperforms state-of-the-art methods on several public 3D face datasets.

\end{abstract}

\section{INTRODUCTION}

Face recognition has been a rapidly developing research task in recent years and has been widely used for many different applications, such as video surveillance, biometric identification, security verification, etc. Although 2D face recognition based on deep learning has achieved very high accuracy in most public datasets, face recognition is still very challenging under large pose variations \cite{abate20072d}. To overcome this problem, some face-frontalization methods have been proposed to normalize profile face images to frontal pose \cite{gao2009pose}\cite{qian2019unsupervised}, and some focused on RGB-D face recognition. Unlike the 2D face recognition approach that uses only RGB images as input, RGB-D face recognition includes depth as additional information, thus leading to more robust performance against large pose and illumination variations. 

The development of 3D or RGB-D face recognition is slower than 2D face recognition. The main reason is the lack of large 3D or RGB-D face datasets available to the public. The numbers of subjects in most 3D or RGB-D face datasets are much smaller than those in 2D face datasets. Numerous 2D datasets contain more than thousands of identities and millions of images \cite{yi2014learning}\cite{cao2018vggface2}\cite{guo2016ms}, whereas existing public 3D face datasets usually contain only hundreds of subjects or at most  thousands of images. \cite{yin20063d}\cite{phillips2005overview}\cite{faltemier2007using}. It is easy to fall into the overfitting problem when we only use a limited number of subjects in a 3D dataset to train a face recognition model. 

To address the problem of lacking large 3D face datasets for model training, many works \cite{zulqarnain2018learning}\cite{kim2017deep}\cite{cai2019fast} applied different data augmentation methods to train their face recognition models. \cite{kim2017deep} changed the values of expression parameters of 3DMM model and randomly generated rigid transformations matrices to the input 3D point cloud to synthesize expression and pose variations. \cite{yi2014learning} generated new identities by morphing two 3D face models of different identities. These methods construct the augmented face data with virtual identity, and it is tough to generate realistic identity-preserving intra-person variations for the synthesized 3D face data for virtual identities.    


\begin{figure}[tb]
    \centering
    \includegraphics[width=0.8\linewidth]{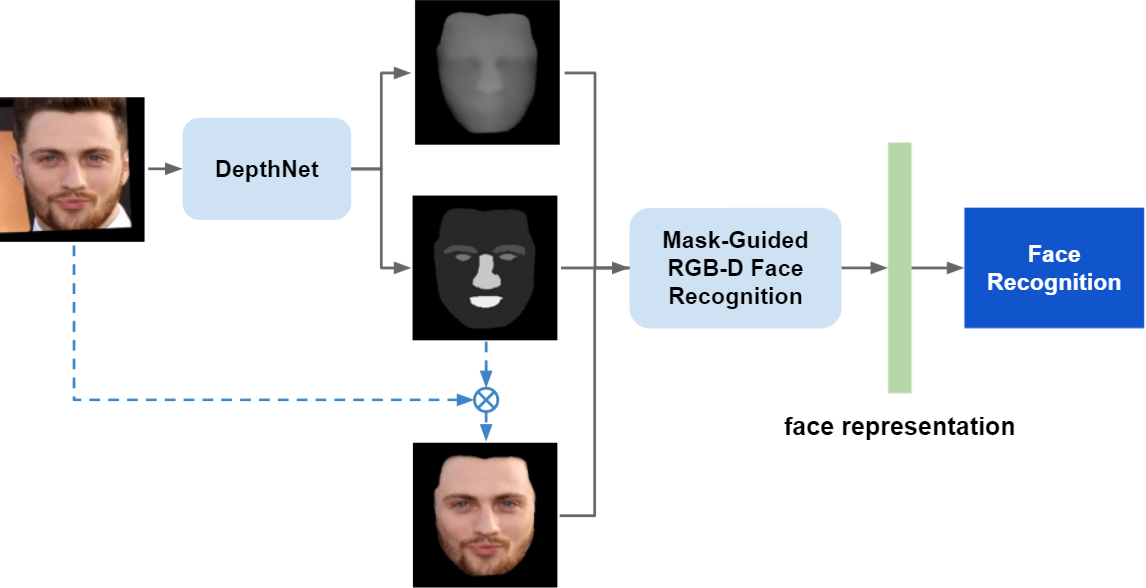}
    \caption{The pipeline of the proposed RGB-D face recognition system}
    \label{fig:pipeline}
\end{figure}

This paper presents a new method to convert a 2D face dataset to a 3D face dataset to address inadequate numbers of subjects in 3D face datasets for model training. Our system has two major parts, i.e., the depth estimation module (DepthNet) and the mask-guided face recognition module. We include a face semantic segmentation branch into the depth estimation network model as an auxiliary task for the depth estimation module. The module can correctly recognize where the facial features are located to estimate realistic face depth images. The proposed mask-guided face recognition model takes RGB face images, segmentation masks, and generated depth maps as input, and this model can achieve high-accuracy RGB-D face recognition. Thus, we can convert a large 2D face dataset to the corresponding RGB-D face dataset with the same number of subjects and intra-variations for training the RGB-D face recognition model. The main contributions of this work can be summarized as follows:
\begin{enumerate}

\item We propose a novel depth estimation CNN model called DepthNet, which includes semantic segmentation to estimate a more accurate depth map than the existing face depth estimation networks. 

\item By applying the proposed DepthNet to a large RGB face image dataset, we obtain the corresponding RGB-D dataset with a large number of subjects and large intra-variations, which can be used for training accurate RGB-D face recognition models.


\item We propose a mask-guided face recognition model which contains  an RGB recognition branch, a depth map recognition branch, and an auxiliary segmentation mask branch with spatial attention module to overcome challenging variations in expression, pose, and occlusion.

\item Experiments on several public 3D face datasets demonstrate that the proposed mask-guided face recognition model outperforms the state-of-the-art methods for RGB-D face recognition.
\end{enumerate}

\section{Related works}


\subsection{3D Data Augmentation}
Due to 3D face data scarcity, many 3D face recognition works focused on developing different 3D data augmentation methods.
Kim et al. \cite{kim2017deep} proposed a 3D face augmentation technique that synthesizes several different facial expressions from a single 3D face scan. Each point cloud from FRGCv2 dataset \cite{phillips2005overview} was fitted to a BFM \cite{paysan20093d} model to produce 25 expressions for each face model by modifying the expression parameters. 
Gilani et al. \cite{zulqarnain2018learning} generated millions of 3D facial models of different virtual identities by simultaneously interpolating between the facial identity and facial expression spaces. 
Zhang et al. \cite{zhang2019data} applied GPMM to generate a large 3D face training dataset and compensated the distribution difference between the generated data and real faces by constraining the face sampling area. 

The methods mentioned above proposed to achieve 3D face data augmentation either via sampling from a low-dimensional identity and expression parametric space for a 3D face morphable model, such as GPMM, or interpolating 3D face models from actual 3D face scans. However, it is still not clear how effective the data synthesis of new virtual identities will benefit the training of face recognition models.  
This paper proposes converting an existing 2D face dataset to an RGB-D face dataset by estimating associated depth maps from 2D face images. A new CNN-based face depth estimation model, called DepthNet, is developed for this specific image-to-image translation problem.

\subsection{RGB-D Face Recognition}
Deep learning-based RGB-D face recognition research is not very active compared to 2D face recognition. One reason is that an effective way to pass the 3D data to the neural network is still under research. Additionally, there is a lack of large 3D face datasets available in public, as mentioned above. Therefore, many works \cite{kim2017deep}\cite{lee2016accurate}\cite{uppal2020attention}\cite{xiong2019improving} employed the CNN that was pre-trained on 2D face images to fine-tune on the relatively small 3D dataset. Gilani et al. \cite{zulqarnain2018learning} took depth, azimuth, and elevation angles of the normal vector as a 3-channel input and proposed the first deep CNN model specifically designed for RGB-D face recognition. Jiang et al. \cite{jiang2019robust} normalized the depth values to the same range as the RGB values and proposed an attribute-aware loss function for CNN-based face recognition to improve the accuracy of recognition results. Li et al. \cite{li2017multimodal} presented a fusion CNN, which took six types of 2D facial attribute maps (i.e., geometry map, three normal maps, curvature map, and texture map) as input for RGB-D facial expression recognition. Instead of using depth data as input, Zhang et al. \cite{zhang2019data} proposed a data-free 3D face recognition method that only used synthesized unreal data from 3D Morphable Model to train a deep point cloud network.


\section{Proposed Method}
We aim to build a robust RGB-D face recognition model from the 2D face image dataset. To achieve this goal, we propose a new CNN model for generating the associated depth map and segmentation mask from an input face image. We can then generate a large RGB-D face dataset from a large 2D RGB face dataset to improve the training of RGB-D face recognition models. Our system consists of two modules: (1) the DepthNet and (2) the mask-guided RGB-D face recognition model. In Fig.~\ref{fig:pipeline}, it is clear to understand the whole process of our method. For each 2D image, we apply FAN face alignment \cite{bulat2017far} as the first step. Second, the augmented depth image and semantic segmentation mask image are generated by the DepthNet. Third, we set the background pixels of the RGB image as zero according to the semantic segmentation mask image. Finally, the face representation is 
computed by a mask-guided RGB-D face recognition model for RGB-D face recognition. Our mask-guided RGB-D face recognition model can also take the acquired depth map as the input by simply replacing the augmented depth image in Fig.~\ref{fig:pipeline} with the actual depth image.

\begin{figure}[tb]
    \centering
    \includegraphics[width=1.0\linewidth]{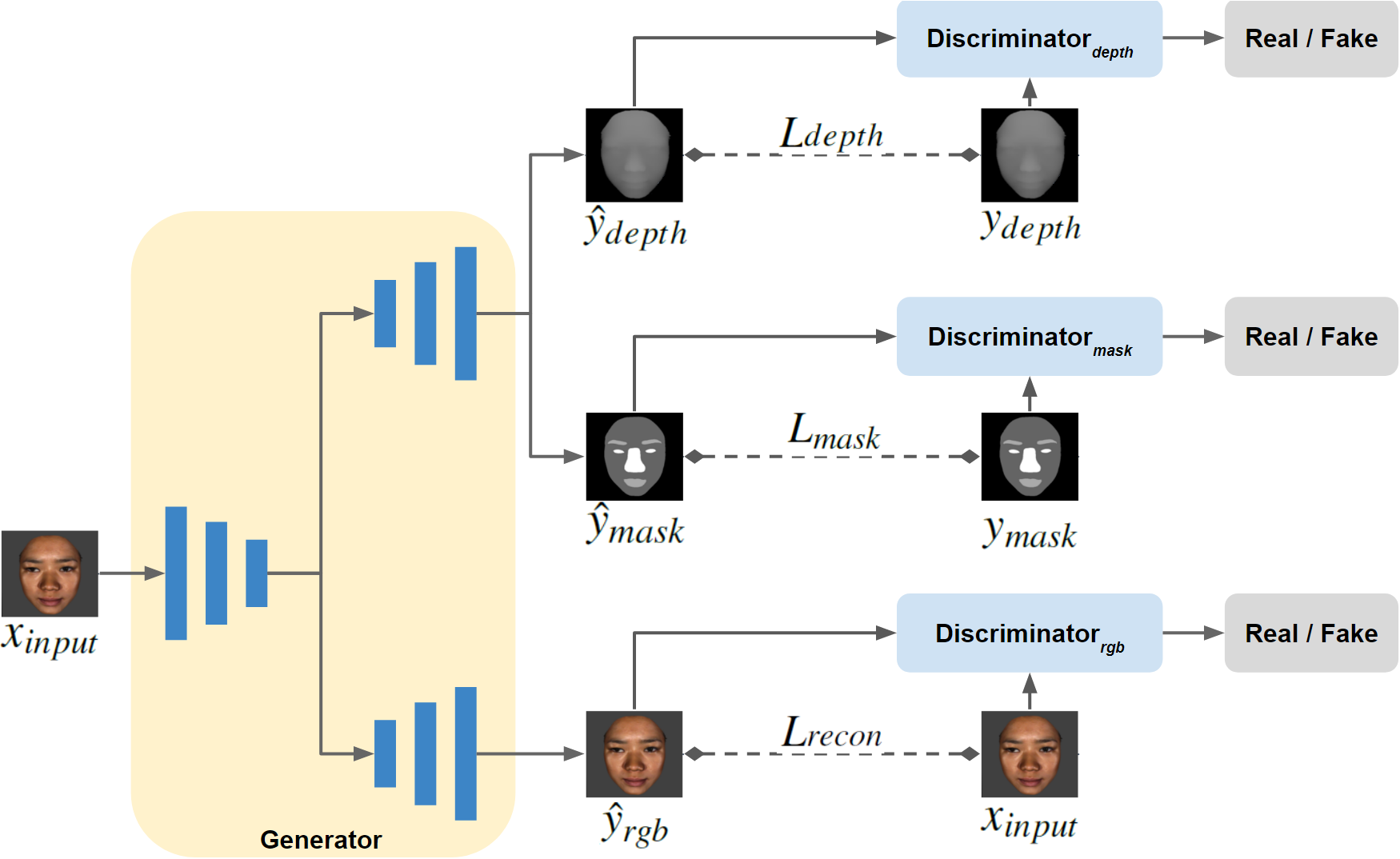}
    \caption{\textbf{Architecture of the proposed DepthNet model.} Each input image ($\hat{X}_{input}$) is first fed into the encoder to encode the face feature vector. Then, the two branches of the decoder generate the semantic segmentation mask ($\hat{y}_{mask}$), the depth map ($\hat{y}_{depth}$), and the reconstructed image ($\hat{y}_{rgb}$) from the embedded features.}
\label{fig:depth_struct}
\end{figure}

\subsection{DepthNet}
Fig.~\ref{fig:depth_struct} illustrates the framework of the proposed DepthNet model, which includes a generator and three discriminators.  The generator can be divided into three networks, the face encoder, the face decoder, and the auxiliary decoder. This generator is based on the UNet \cite{ronneberger2015u} architecture, which is an encoder-decoder model with a skip-connection module. With the skip-connection module, the decoder can directly use the features from the encoder. For a given source face image $X_{input}$, which passes through the face encoder and the auxiliary decoder to encode image $x_{input}$ information. 

We obtain the reconstructed image $\hat{y}_{rgb}$ with the face decoder. To minimize the distance between the reconstructed image $\hat{y}_{rgb}$ and source face image $X_{input}$, we adopt the L1 loss as follows: 
\begin{equation}
L_{recon}= \mathbb{E}_{x\sim P_{x}}[\left \| x_{input}-\hat{y}_{rgb} \right \|_{1}]
\end{equation}

\begin{figure*}[t]
    \centering
    \includegraphics[width=1.0\linewidth]{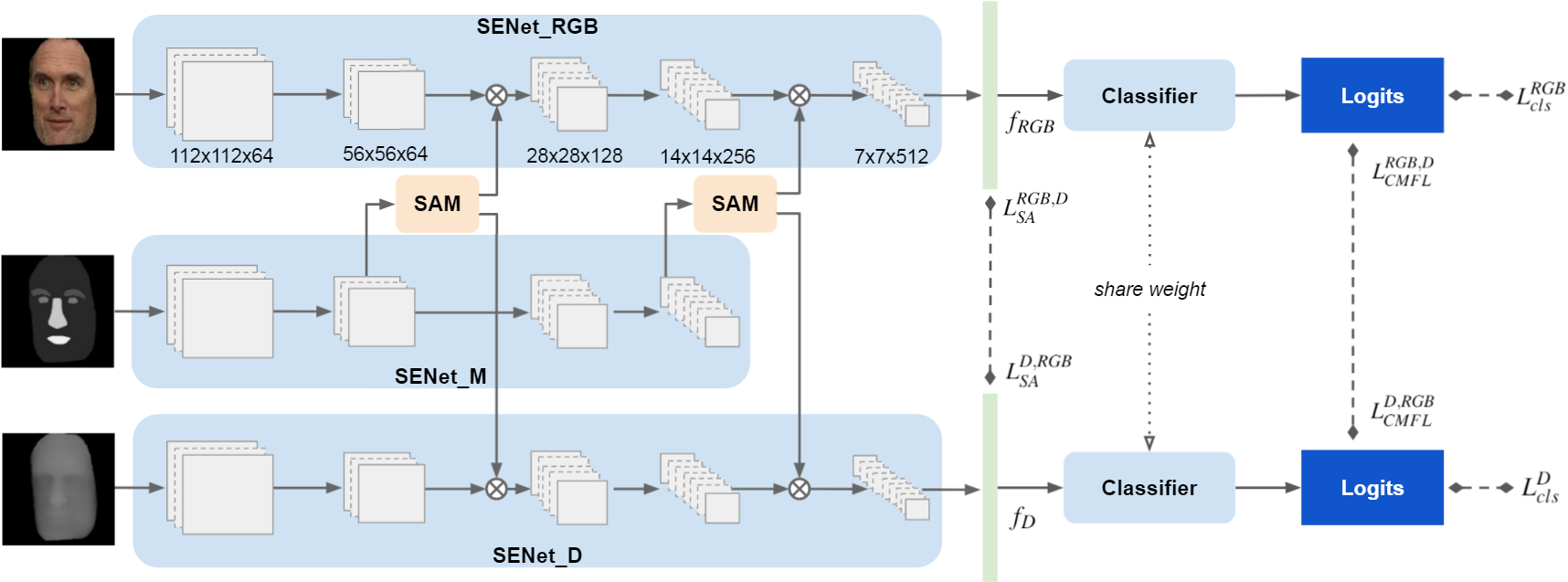}
    \caption{The proposed mask-guided RGB-D face recognition network architecture.}
    \label{fig:recognition}
\end{figure*}

Meanwhile, the auxiliary decoder generates the corresponding depth map $\hat{y}_{depth}$ and semantic segmentation mask $\hat{y}_{mask}$ of the input face image $X_{input}$. We design a shared weight architecture to output the segmentation mask and depth at the same time. To minimize the distance between the generated depth image $\hat{y}_{depth}$ and ground truth depth $y_{depth}$, we adopt the L1 loss as follows:
\begin{equation}
\begin{split}
L_{depth}= \mathbb{E}_{x\sim P_{x}}[\left \| y_{depth}-\hat{y}_{depth} \right \|_{1}] 
\end{split}
\end{equation}
We adopt the binary cross-entropy loss to train the network to generate the semantic segmentation mask for an input image. This loss enforces the output of the encoder to be similar to the ground-truth semantic segmentation. It is given by 
\begin{equation}
\begin{split}
L_{mask} = &\mathbb{E}_{x\sim P_{x}} -(y_{mask}\cdot log(\hat{y}_{mask})\\
&+(1-y_{mask})\cdot log(1-\hat{y}_{mask}))
\end{split}
\end{equation}
where $\hat{y}_{mask}$ denotes the generated segmentation mask for the input face image, and $y_{mask}$ is the ground truth segmentation mask. We also leverage generative models to learn to reconstruct images to train the RGB discriminator $D_{rgb}$, depth discriminator $D_{d}$, and mask discriminator $D_{m}$, given by
\begin{equation}
\begin{split}
L_{adv}^{Gen} = &\mathbb{E}_{y\sim P_{y}} [(D_{rgb}(\hat{y}_{rgb})-1)^2]\\
&+\mathbb{E}_{y\sim P_{y}} [(D_{d}(\hat{y}_{depth})-1)^2]\\
&+\mathbb{E}_{y\sim P_{y}} [(D_{m}(\hat{y}_{mask})-1)^2]
\end{split}
\end{equation}
\begin{equation}
\begin{split}
L_{adv}^{Dis} = & \mathbb{E}_{y\sim P_{y}} [(D_{rgb}(x_{input}) - 1)^2]\\
& +\mathbb{E}_{x\sim P_{x}}[(D_{rgb}(\hat{y}_{rgb}))^2]\\
& +\mathbb{E}_{y\sim P_{y}} [(D_{d}(y_{depth}) - 1)^2]\\
&+\mathbb{E}_{x\sim P_{x}}[(D_{d}(\hat{y}_{depth}))^2]\\
& +\mathbb{E}_{y\sim P_{y}} [(D_{m}(y_{mask}) - 1)^2]\\
&+\mathbb{E}_{x\sim P_{x}}[(D_{m}(\hat{y}_{mask}))^2]
\end{split}
\end{equation}

The overall loss function for training the DepthNet is given as follows:
\begin{equation}
\begin{split}
L_{Total} = & L_{G}+L_{adv}^{Dis}\\
\end{split}
\end{equation}
\begin{equation}
\label{eq:7}
L_{G} =  \lambda_{1}L_{depth}+\lambda_{2}L_{mask}
+\lambda_{3}L_{adv}^{Gen}+L_{recon}\\
\end{equation}
where $\lambda_{1}$, $\lambda_{2}$ and $\lambda_{3}$ are the weights used to balance the three loss terms.

\subsection{Mask-Guided RGB-D Face Recognition}
Fig.~\ref{fig:recognition} demonstrates our mask-guided RGB-D face recognition network architecture, which contains an RGB recognition branch, a depth map recognition branch, and an auxiliary segmentation mask branch with spatial attention module proposed in \cite{Woo_2018_ECCV}. At the training stage, the RGB recognition branch extracts the face representation feature, $f_{RGB}$ in the figure, by the backbone network SENet \cite{hu2018squeeze} network denoted as SENet\_RGB. Similarly, the depth map recognition branch extracts corresponding $f_D$ by SENet\_D. The auxiliary segmentation mask branch extracts different level of feature map from segmentation mask by SENet\_M, and then applies spatial attention module (SAM) on those feature maps to aid RGB and D branches while training. This SAM is shared-weighted across the RGB recognition branch and D recognition branch. It can provide auxiliary information from the segmentation branch to help recognition branches focus on the informative parts on segmentation feature maps. Finally, the classifier with ArcFace \cite{deng2019arcface} additive angular margin loss predicts a vector of probabilities with one value for each possible identity.

The proposed mask-guided RGB-D face recognition network is a two-stream-multi-head architecture, and we apply the cross-entropy loss as classification losses $L_{cls}$ on individual branches. We adopt the cross-modal focal loss $L_{CMFL}^{m, n}$ in \cite{cmfl} to learn robust representations jointly, which is defined as
\begin{equation}
\label{eq:8}
\begin{split}
    L_{CMFL}^{m, n} = -\alpha(1-w(m_t, n_t))^{\gamma} \log(m_t)
\end{split}
\end{equation}
$\alpha$ and $\gamma$ are tunable hyper-parameters.
\begin{equation}
\begin{split}
    w(m_t, n_t) = n_t\frac{2m_tn_t}{m_t+n_t}
\end{split}
\end{equation}
where $m_t$ and $n_t$ denote the classification probabilities after fully connected layer in current branch $m$ and the other branch $n$, respectively. The CMFL contributed by branch $n$ will reduce when branch $n$ can predict with high confidence.

Although the inputs to the mask-guided RGB-D face recognition network could be a different combination of modalities, their inputs should represent the same semantic meaning as the subject identity. Inspired by \cite{Abavisani_2019_CVPR}, we add another semantic alignment loss $L_{SA}^{m,n}$ to share semantics for the extracted feature vectors $f_{m}$ and $f_n$, given by
\begin{equation}
\begin{split}
    L_{SA}^{m, n} = \rho^{m,n}(1-\text{cosine\_similarity}(f_{m}, f_{n}))
\end{split}
\end{equation}
where $\rho^{m,n}$ is the focal regularization parameter to make sure the network will only transfer information from the more accurate network to the weaker network. For current modality $m$ and the other modality $n$, we use the difference of classification losses between $m$ and $n$ to measure the performance of the network, and it is denoted as $L_{cls}^{m}-L_{cls}^{n}$. If the difference is positive, then it means modality $m$ is weaker than modality $n$. The model will enforce $f_{m}$ to be similar to $f_{n}$. The focal regularization parameter is defined as follows
\begin{equation}
\label{eq:11}
\begin{split}
    \rho^{m,n}
    =\begin{cases}
        e^{\beta (L_{cls}^{m}-L_{cls}^{n})}-1, & \text{if $L_{cls}^{m} > L_{cls}^{n}$}\\
        0, & \text{if $L_{cls}^{m} \le L_{cls}^{n}$}
    \end{cases}
\end{split}
\end{equation}
where $\beta$ is a positive focusing parameter.

The overall loss functions in branch RGB and D are given as
\begin{equation}
\label{eq:12}
\begin{split}
    L_{total}^{RGB} = (1-\lambda_1)L_{cls}^{RGB} + \lambda_1L_{CMFL}^{RGB,D} + \lambda_2L_{SA}^{RGB,D}
\end{split}
\end{equation}
\begin{equation}
\label{eq:13}
\begin{split}
    L_{total}^{D} = (1-\lambda_1)L_{cls}^{D} + \lambda_1L_{CMFL}^{D,RGB} + \lambda_2L_{SA}^{D,RGB}
\end{split}
\end{equation}
The total loss of RGB branch $L_{total}^{RGB}$ optimizes the parameters of RGB recognition branch and auxiliary segmentation branch. Similarly, the total loss of D branch $L_{total}^{D}$ optimizes the parameters of the D recognition branch and auxiliary segmentation branch.

At the testing stage, instead of simply concatenating $f_{RGB}$ and $f_D$, we compute two cosine similarity scores and perform score-level fusion by averaging two cosine similarity scores to give the final prediction. Our experimental result shows that the combination of all modalities provides the most accurate result for face recognition. More details will be provided in the next section.

\section{Experimental Results}

\subsection{Experimental Datasets}
Here we introduce the datasets that were used in the training and evaluation. The DepthNet model is trained with BU-3DFE 3D database \cite{bu3dfe} dataset. The VGGFace2 \cite{cao2018vggface2} 2D face dataset is augmented to the corresponding RGB-D face dataset by applying the DepthNet model, and then is used to train the mask-guided RGB-D face recognition model. We experiment on public 3D datasets: BU-3DFE \cite{bu3dfe}, Texas FR3D \cite{texas}, Bosphorus \cite{bosphorus}, FRGCv2 \cite{phillips2005overview}, and Lock3DFace \cite{lock3dface} to evaluate the proposed DepthNet and mask-guided RGB-D face recognition model.

\subsubsection{Training Data Preparation}
Since we aim to make DepthNet learn how to convert an RGB face image to the corresponding depth image, we adopt 90 identities from BU-3DFE 3D face database\cite{bu3dfe} as the training data. Also, the pose augmentation is implemented by rotating the original frontal face point cloud along with the yaw and pitch axis. 
Next, We modify the BiSeNet\cite{bisent} model to generate the semantic segmentation mask for an input face image and take the results as the pseudo ground truth segmentation mask. 
The segmentation mask consists of seven channels representing different labels: background, skin, brows, eyes, glasses, nose, and mouth. Then, we use these RGB-D images and the corresponding pseudo-ground-truth segmentation masks to train our DepthNet. 

For RGB-D face recognition, we select VGGFace2 \cite{cao2018vggface2} as our training data. It contains 9,131 subjects and a total of 3.31 million RGB images. The proposed DepthNet model produces the corresponding depth images and segmentation masks. The augmented depth images will be gray images with a channel equal to one and a seven-channel image representing the segmentation mask.  We can generate an even larger RGB-D dataset for model training by using larger RGB datasets. However, due to the memory and training time consideration, we choose to use the VGGFace2 dataset for conversion into the RGB-D dataset to train our face recognition model.

\subsubsection{3D Face Datasets}\hfill\\
{\bf BU-3DFE 3D database}\cite{bu3dfe} includes 100 subjects with 2,500 scans. Each identity performs seven expressions with four levels of intensity for each expression except for the neutral one. There is no pose variation in this database. The evaluation protocol as \cite{FR3DNet} is adopted so that we have 100 images in the gallery and 2,400 images in the probe to calculate the identification accuracy.

{\bf Texas FR3D database}\cite{texas} contains 1,149 scans of 118 subjects. All the scans are frontal with different expressions. We select the first images for all the 118 subjects as the gallery and put the remaining 1,031 images as the probe.

{\bf Bosphorus database}\cite{bosphorus} consists of 4,666 scans of 105 subjects in various poses, expressions and occlusions. There are two settings for evaluation. {\it Setting-1} considers expression variations for recognition. The first neutral image of each subject is selected as the gallery. The other 2,797 images are regarded as probe images. {\it Setting-2} takes all the remaining 4,561 images in the probe and 105 images in the gallery.

{\bf FRGCv2 database}\cite{phillips2005overview} contains images from 466 subjects collected in 4,007 scans with two facial expression variances (e.g., neutral and smile). We select the first neutral images from all the 466 subjects as the gallery and take the remaining 3,541 images as the probe.

{\bf BUAA Lock3DFace database}\cite{lock3dface} contains 5711 RGB-D face videos of 509 subjects with variations in facial expression, pose, occlusion and time-lapse. We follow the same testing protocols as described in \cite{lock3dface}. The first neutral image of each subject in Session-1 (S-1) is selected as the gallery. And then divide the remaining into four test sets: Probe\_Set\_1 for images with expression changes in S-1; Probe\_Set\_2 for images with pose variations in S-1; Probe\_Set\_3 for images with occlusions in S-1; and Probe\_Set\_4 for all images in Session-2 (S-2).

\subsection{Implementation Details}
For training DepthNet, we adopt Adam as the optimizer with setting $\beta_1 = 0.5$, $\beta_2 = 0.999$ and learning rate $\gamma$ is set to 0.0002. For the hyper-parameter of the loss function in~(\ref{eq:7}), we set $\lambda_1 = 100$, $\lambda_2 = 100$ and $\lambda_3 = 1$. We train the DepthNet model on a GTX1080Ti GPU card with batch size equal to 16 and image size 256x256.

When training the RGB-D face recognition model, we use SGD as the optimizer with momentum = 0.9, weight decay = 0.0005 and learning rate = 0.1 divided by 10 at 6, 10, 17 epochs. The SAM is applied on feature map with size of 56x56x64 and 14x14x256. We set $\alpha=1$ and $\gamma=3$ in~(\ref{eq:8}) and set $\beta=2$ in~(\ref{eq:11}). For the hyper-parameter of the loss function in~(\ref{eq:12}) and~(\ref{eq:13}),we set $\lambda_1 = 0.5$ and $\lambda_2 = 0.05$. We train RGB-D face recognition on 2 Tesla V100 GPUs with batch size 256 and image size 112x112.

\subsection{DepthNet Evaluation} \label{sec:4.3}
\begin{figure}[t]
    \centering
    \includegraphics[width=1.0\linewidth]{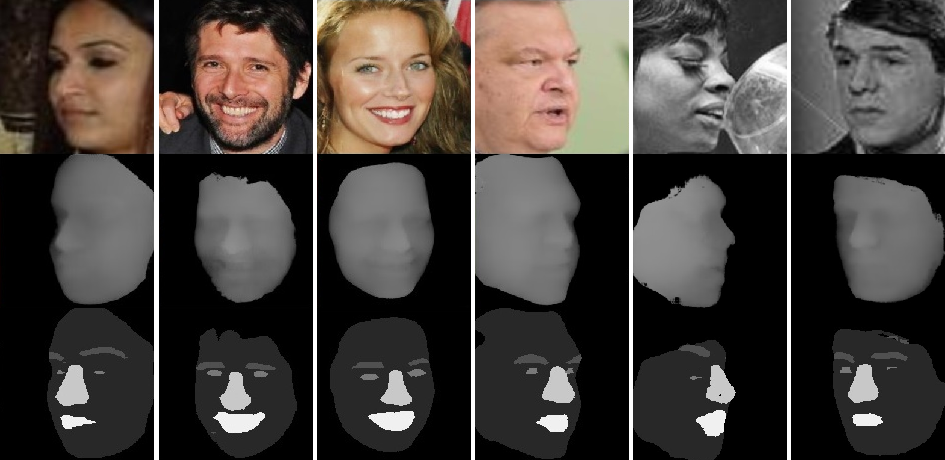}
    \caption{Generated augmented depth and segmentation images of VGGFace2. Rows from top to bottom: RGB images, augmented depth images, and segmentation images.}
\label{fig:vggdm}
\end{figure}

In this section, we demonstrate some results of our proposed depth estimation method. Our proposed DepthNet aims to produce additional augmented depth images and segmentation mask images for 2D datasets. As a result, in Fig.~\ref{fig:vggdm},  we depict some examples of applying our DepthNet to VGGFace2 2D face database, which is the training set for our RGB-D face recognition model. The results show that our method can produce well-preserved face contour and face shape features of different expressions.

\begin{table}[t]
\caption{Quantitative comparison of depth estimation errors for different methods: MSE between ground truth depth and estimated depth images.}
\begin{center}
\begin{tabular}{|l|ccc|}
\hline
Method & BU3DFE & FRGCv2 & Bosphorus\\
\hline\hline
3DDFA \cite{zhu2017face} & 125.78 & 597.17 & 540.48 \\
PRNet \cite{feng2018joint} & 74.86 & \bf{216.29} & 615.99\\
\hline
Ours & \bf{16.65} & 435.88 & \bf{421.46}\\
\hline
\end{tabular}
\end{center}
\label{tab:depth_mse}
\end{table}

\begin{figure}[tb]
    \centering
    \includegraphics[width=0.95\linewidth]{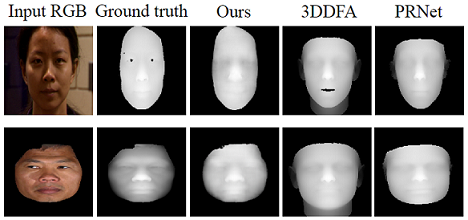}
    \caption{Depth estimation results by using different methods on some sample images in FRGCv2 dataset (top row) and Texas dataset (bottom row).}
    \label{fig:frgcv2_depth}
\end{figure}

In Table~\ref{tab:depth_mse}, we compute Mean Square Error (MSE) between the generated depth images and ground truth depth images with comparison with two 3D face depth estimation methods, 3DDFA \cite{3ddfa} and PRNet \cite{prnet}. 
For a fair comparison, we only calculate the MSE in the intersection of ground truth depth and all predicted depth images from our DepthNet, 3DDFA, and PRNet. 
For the BU3DFE database, it is evident that we have the best performance on the testing set of BU3DFE partially because we train DepthNet with the training set of BU3DFE, which leads to negligible bias between training and testing data. For FRGCv2 dataset, although the estimation results by our model are not the best among the three methods, our DepthNet model can generate a more accurate depth image around the face contour than the other two methods, as shown in  Fig.~\ref{fig:frgcv2_depth}. This is because our model includes semantic segmentation together with depth estimation. Our DepthNet is trained with BU3DFE which was acquired with a structured-light-based  3D sensor. The Bosphorus 3D images were also acquired using a structured-light-based device. However, the FRGCv2 3D images were captured by a laser-based sensor. As a result, the improvement is not as significant as the others. Our DepthNet achieves the best performance on the Bosphorus dataset, which contains large pose variations, and our DepthNet was trained with such variations.

\begin{figure}[tb]
    \centering
    \includegraphics[width=1.0\linewidth]{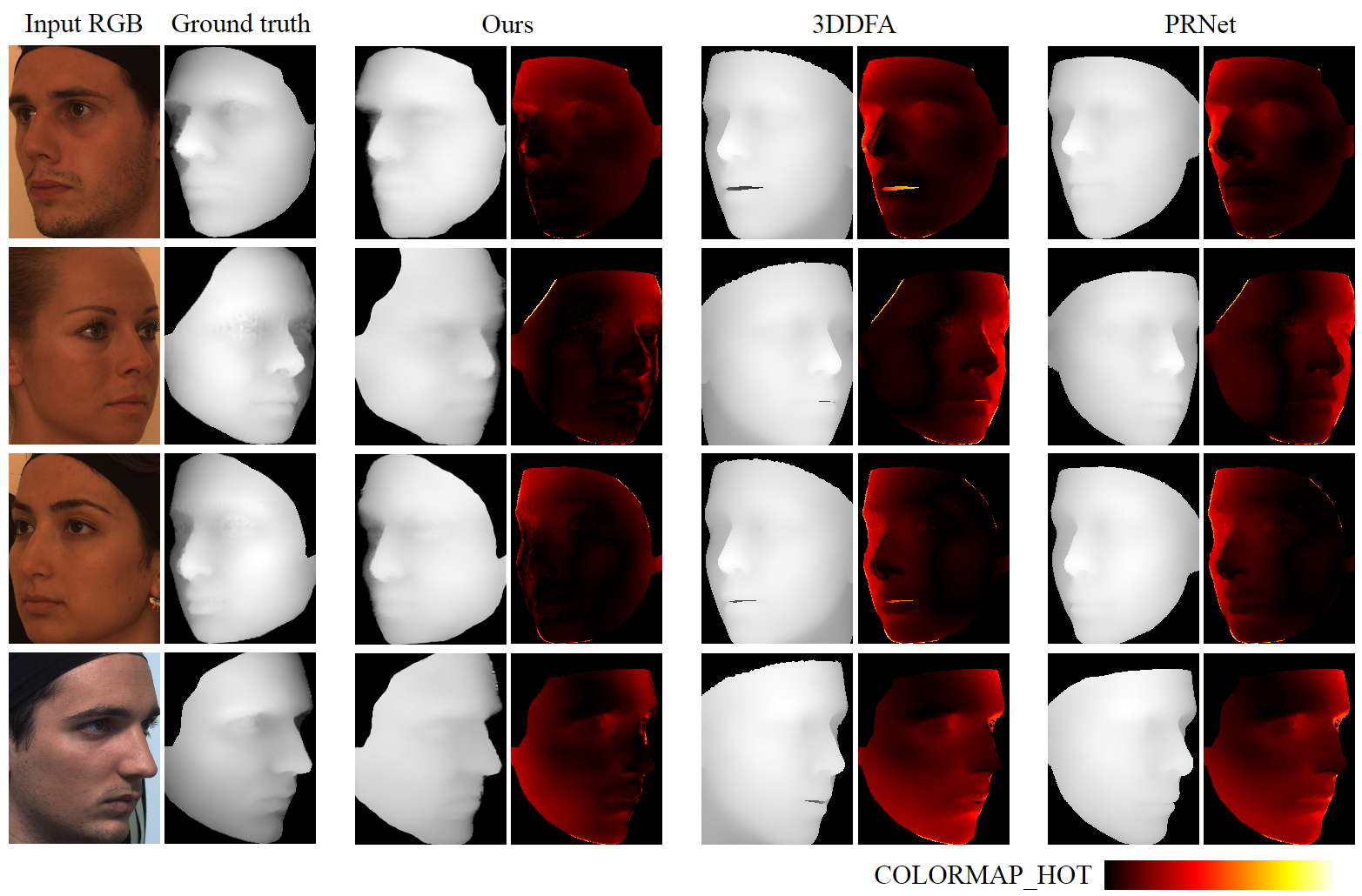}
    \caption{Depth estimation results by using different methods on the Bosphorus dataset. We utilize hot colormap to illustrate the MSE. The darker the color, the smaller the error.}
    \label{fig:bosphorus_map}
\end{figure}

In Figure~\ref{fig:bosphorus_map}, we further illustrate how our DepthNet provides superior depth estimation for face images with large poses. The other two methods have large deviations near the face profile regions. With an additional semantic segmentation branch, our DepthNet can recognize the facial regions from the image to generate an accurate and plausible depth map that is consistent with the RGB face image.

\begin{table}[t]
\caption{The rank-1 identification accuracy on public 3D face databases.}
\begin{center}
\begin{tabular}{|l|cccc|}
\hline
Method & BU3DFE & Texas & Bosphorus-1 & Bosphorus-2\\
\hline\hline
Li {\it et al.}\cite{li2015towards} & - & - & 98.8 & 96.6\\
Lei {\it et al.}\cite{lei2016two} & 93.25 & - & 98.9 & -\\
Mian {\it et al.}\cite{mian2007efficient} & 95.9 & 98.0 & - & 96.4\\
Lin {\it et al.}\cite{lin2019local} & 96.2 & - & 99.71 & -\\
Kim {\it et al.}\cite{kim2017deep} & 95.0 & - & 99.2 & -\\
FR3DNet \cite{zulqarnain2018learning} & 98.64 & 100 & - & 96.18\\
\hline
Ours & \bf{100} & \bf{100} & \bf{100} & \bf{97.94}\\
\hline
\end{tabular}
\end{center}
\label{tab:accuracy_RGBDM}
\end{table}
\begin{table*}[t]
\caption{Rank-1 identification accuracy applied to different modalities. VGGFace2$^{*}$ denotes the augmented data of VGGFace2 that produced by DepthNet. D$^{*}$ and M$^{*}$ denotes the augmented depth map and segmentation mask generated by DepthNet.}
\begin{center}
\begin{tabular}{|l|ccc|cc|}
\hline
Method & Training data & Subjects & Testing Modality & FRGCv2 & Bosphorus-1\\
\hline\hline
VGG-Face\cite{cao2018vggface2}& Private\cite{zulqarnain2018learning} & 100 & RGB & 87.92 & 96.39\\
Jiang {\it et al.}\cite{jiang2019robust} & TRAINING-SET-I\cite{jiang2019robust} & 60,000 & RGB & 95.69 & 96.08\\
Ours & VGGFace2$^{*}$\cite{cao2018vggface2} & 9,131 & RGB + M$^{*}$ & \bf{99.07} & \bf{99.75}\\ 
\hline
Li {\it et al.}\cite{li2015towards} & Part of Bosphorus\cite{bosphorus} & 105 & Depth & 96.30 & 95.40\\
FR3DNet\cite{zulqarnain2018learning} & Private\cite{zulqarnain2018learning} & 100 & Depth & 97.06 & 96.18\\
Jiang {\it et al.}\cite{jiang2019robust} & TRAINING-SET-I\cite{jiang2019robust} & 60,000 & Depth & 97.45 & \bf{99.37}\\
Ours & VGGFace2$^{*}$\cite{cao2018vggface2} & 9,131 & D$^{*}$ + M$^{*}$& \bf{98.42} & 98.61\\
\hline
Li {\it et al.}\cite{li2015towards} & Part of FRGCv2\cite{phillips2005overview} & 466 & RGB + Depth & 95.20 & 99.40\\
Jiang {\it et al.}\cite{jiang2019robust} & TRAINING-SET-I\cite{jiang2019robust} & 60,000 & RGB + Depth & 98.52 & 99.52\\ 
Ours & VGGFace2$^{*}$\cite{cao2018vggface2} & 9,131 & RGB + D$^{*}$ + M$^{*}$& \bf{99.27} & \bf{100}\\
\hline
\end{tabular}
\end{center}
\label{tab:accuracy_Modality}
\end{table*}
\begin{table*}[t]
\caption{The rank-1 identification accuracy on Lock3DFace databases. D$^{*}$ and M$^{*}$ denotes the augmented depth map and segmentation mask generated by DepthNet.}
\begin{center}
\begin{tabular}{|l|l|ccccc|}
\hline
 & & & & Accuracy & & \\
Method & Input & Expression & Pose & Occlusion & Time & Average\\
\hline\hline
He {\it et al.}\cite{7780459}  & RGB & 96.3 & 58.4 & 74.7 & 75.5 & 76.2\\
Hu {\it et al.}\cite{hu2018squeeze}  & RGB & 98.2 & 60.7 & 77.9 & 78.3 & 78.7\\
Cui {\it et al.}\cite{8411215} & RGB + D & 97.3 & 54.6 & 69.6 & 66.1 & 71.9\\
Mu {\it et al.}\cite{8953221}  & RGB + 3D Model & 98.2 & 70.4 & 78.1 & 65.3 & 84.2\\
Uppal {\it et al.}\cite{9330625} & RGB + D & 99.4 & 70.6 & 85.8 & 81.1 & 87.3\\
\hline
Ours & RGB + D$^*$ + M$^*$ & \bf{99.92} & \bf{96.55} & \bf{97.31} & \bf{92.38} & \bf{96.43}\\
\hline
\end{tabular}
\end{center}
\label{tab:lock3dface}
\end{table*}
\begin{table}[t]
\caption{The comparison of MSEs of depth estimation with and without including the semantic segmentation task.}
\begin{center}
\begin{tabular}{|l|ccc|}
\hline
Method & BU3DFE & FRGCv2 & Bosphorus\\
\hline\hline
DepthNet {\it w/o} mask & 84.62 & 880.99 & 848.88\\
DepthNet {\it w} mask & \bf{42.66} & \bf{605.48} & \bf{839.86}\\
\hline
\end{tabular}
\end{center}
\label{tab:mask_mse}
\end{table}
\begin{table}[t]
\caption{Identification accuracy on Lock3DFace dataset of our mask-guided RGB-D face recognition model and its variants with different modalities.}
\begin{center}
\begin{tabular}{|l|ccccc|}
\hline
& & & Lock3DFace & & \\
Method & Expression & Pose & Occlusion & Time & Average\\
\hline\hline
D$^{*}$+M$^{*}$ & 99.46 & 81.36 & 79.08 & 71.89 & 83.12\\
RGB+M$^{*}$ & 99.77 & 93.98 & 94.22 & 88.68 & 94.09\\
RGB+D$^{*}$+M$^{*}$ & \bf{99.92} & \bf{96.55} & \bf{97.31} & \bf{92.38} & \bf{96.43}\\
\hline
\end{tabular}
\end{center}
\label{tab:accuracy_DepthNet}
\end{table}

\subsection{Mask-Guided RGB-D Face Recognition Evaluation}
Our mask-guided RGB-D face recognition model has good generalization ability on other 3D datasets, even though it is trained with a depth-augmented 2D dataset. Our network is trained on VGGFace2 \cite{cao2018vggface2} and directly tested on all other 3D face datasets without any fine-tuning. Our mask-guided RGB-D face recognition model takes RGB face image, augmented depth image generated by DepthNet (D$^*$), and augmented segmentation mask generated by DepthNet (M$^*$) as input. We demonstrate the rank-1 identification results on some public 3D face databases in table~\ref{tab:accuracy_RGBDM}. For all the datasets, our method provides state-of-the-art verification accuracy. Especially for Bosphorus-2, which has pose variations, the proposed method marginally outperforms the other methods by around 1.7\% accuracy. 

Table~\ref{tab:accuracy_Modality} further shows that our model can also be applied to different modalities. We compare our results with other RGB-D face recognition methods and report the rank-1 identification accuracy on FRGCv2 and Bosphorus-1. Our mask-guided recognition model trains the RGB and D branches jointly; the training data that includes the augmented depth or segmentation mask images of VGGFace2 are denoted as VGGFace2$^{*}$. Our DepthNet can effectively transform a 2D face image into the corresponding RGB-D image to resolve the problem that the existing public 3D face database usually has inadequate subjects or intra-person variations. Jiang {\it et al.}\cite{jiang2019robust} proposed an attribute-aware loss function and a newly collected RGB-D face database with 60K subjects to improve the accuracy of RGB-D face recognition results. We can observe that our proposed method, trained with the RGB-D dataset with augmented depth, segmentation masks, and 9K subjects, is superior to the model trained with ground truth depth images and many more subjects. With the proposed method, we can get an RGB-D database with sufficient subjects from the existing 2D face databases and do not need to collect a new 3D face database.

The rank-1 identification accuracies for the Lock3DFace dataset are shown in Table~\ref{tab:lock3dface}. Especially in the subset with pose variations, our result achieves a 96.55\% accuracy which is significantly better (+25\%) than others. For occlusion variations such as covering the face with hand or glasses, we reach an accuracy of 97.31\% obtaining about +12\% performance gain. In the subset over time scenario, our method also accomplished an accuracy of 92.38\%, which exceeds others by +11\%. It is worth noting that some other methods include part of the Lock3DFace in their training set; However, our mask-guided directly test on Lock3DFace without any fine-tuning. In general, our mask-guided RGB-D recognition model achieves a much higher (+9\%) average accuracy of 96.43\% comparing to other state-of-the-art methods. This indicates that our mask-guided FR model fully exploits the augmented depth and segmentation mask information and is more robust against pose variation than other RGB-D face recognition methods. 

\begin{figure}[t]
    \centering
    \includegraphics[width=1.0\linewidth]{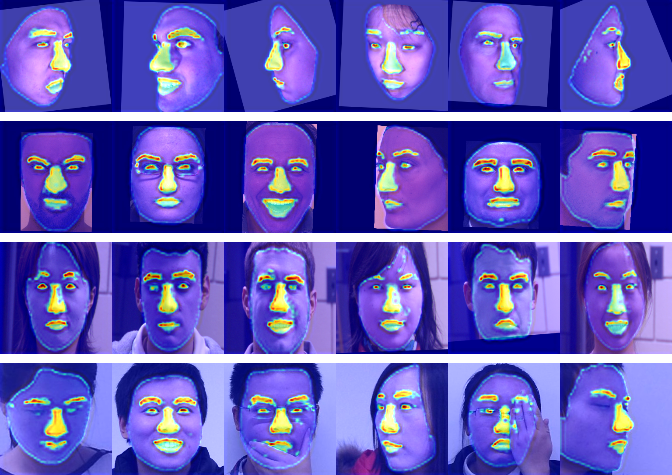}
    \caption{Spatial attention maps four RGB-D face datasets. First row: BU-3DFE dataset; Second row: Bosphorus datase; Third row: FRGCv2 dataset; and Last row: Lock3DFace dataset.}
    \label{fig:vis_attmap}
\end{figure}

Fig.~\ref{fig:vis_attmap} demonstrates visualization results of the mask-guided spatial attention module on some pulic 3D datasets. The result shows some samples with expression, pose, and occlusion variations. The segmentation mask branch provides auxiliary information to spatial attention module; therefore, we can observe that the attention have selectively focused on the informative parts such as eyes, nose, eyebrows, and lips for RGB-D face recognition.

\section{Ablation Study}

\subsection{Effect of the Segmentation Mask}
In this section, we first analyze the effects of the segmentation mask branch in the proposed DepthNet. Table~\ref{tab:mask_mse} demonstrates significant improvement of the depth estimation by including the semantic segmentation into the model. Different from section~\ref{sec:4.3}, we directly calculate the MSE between the estimated depth image and the ground truth image. We can observe that 
with the addition of the semantic segmentation branch, it can focus on the face features and provide precise depth estimation. We can easily perceive the expression of both profile images and frontal images with the segmentation mask.

\subsection{Effect of the DepthNet}
We report the rank-1 face identification results of the proposed mask-guided RGB-D face recognition module and its variants with different combination of modalities, RGB images or augmented depth images D$^{*}$ or segmentation mask images M$^{*}$, as the ablation study. The comparison results are presented in Table~\ref{tab:accuracy_DepthNet}. The first row is the result that we test with augmented depth and the augmented segmentation mask images. The second row is tested with RGB and the augmented segmentation mask images. The third row is tested with RGB, the augmented depth, and the augmented segmentation mask images. We can observe that using both augmented depth and segmentation mask images achieves the highest accuracy, which indicates each component is essential in our mask-guided RGB-D face recognition method.


\section{CONCLUSIONS}

In this paper, we propose a novel framework that estimates depth maps from RGB face images by including a semantic segmentation module for more precise face region localization. The estimated depth maps can be combined with 2D images to augment the 2D face image dataset to RGB-D face dataset. This data augmentation approach helps to improve the accuracy and stability for training the RGB-D face recognition model. Furthermore, we developed a mask-guided RGB-D face recognition model, which includes the auxiliary segmentation attention module to fully exploit the augmented depth and segmentation mask information. Our experiments showed that our DepthNet model provide accurate depth map estimation and the proposed mask-guided RGB-D face recognition model outperforms state-of-the-art face recognition methods on several public 3D face datasets.

{\small
\bibliographystyle{ieee}
\bibliography{egbib}
}

\end{document}